\pdfoutput=1
\PassOptionsToPackage{dvipsnames}{xcolor}
\documentclass[11pt]{article}
\usepackage[table]{xcolor}
\usepackage{ACL2023}

\usepackage{graphicx} 
\graphicspath{ {./Images/} }
\usepackage{comment}
\usepackage{hyperref}
\usepackage{amsmath}
\usepackage{amssymb}
\usepackage{makecell}
\usepackage{transparent}
\usepackage{booktabs}
\usepackage{multicol}
\usepackage{multirow}

\usepackage{times}
\usepackage{latexsym}

\usepackage[T1]{fontenc}

\usepackage[utf8]{inputenc}

\usepackage{microtype}

\usepackage{inconsolata}

\title{Revealing the Blind Spot of Sentence Encoder Evaluation by H\textsc{eros}}

\author{Cheng-Han Chiang$^\dagger$ \hspace{1mm}  Yung-Sung Chuang$^\ddagger$ \hspace{1mm} James Glass$^\ddagger$ \hspace{1mm} Hung-yi Lee$^\dagger$\\
\texttt{dcml0714@gmail.com} \hspace{1mm} \texttt{yungsung@mit.edu} \\ 
   National Taiwan University$^\dagger$ \quad
  Massachusetts Institute of Technology$^\ddagger$ \\}

\begin{document}
\maketitle
\begin{abstract}
Existing sentence textual similarity benchmark datasets only use a single number to summarize how similar the sentence encoder's decision is to humans'.
However, it is unclear what kind of sentence pairs a sentence encoder (SE) would consider similar.
Moreover, existing SE benchmarks mainly consider sentence pairs with low lexical overlap, so it is unclear how the SEs behave when two sentences have high lexical overlap.
We introduce a high-quality SE diagnostic dataset, H\textsc{eros}.
H\textsc{eros} is constructed by transforming an original sentence into a new sentence based on certain rules to form a \textit{minimal pair}, and the minimal pair has high lexical overlaps.
The rules include replacing a word with a synonym, an antonym, a typo, a random word, and converting the original sentence into its negation.
Different rules yield different subsets of H\textsc{eros}.
By systematically comparing the performance of over 60 supervised and unsupervised SEs on H\textsc{eros}, we reveal that most unsupervised sentence encoders are insensitive to negation.
We find the datasets used to train the SE are the main determinants of what kind of sentence pairs an SE considers similar.
We also show that even if two SEs have similar performance on STS benchmarks, they can have very different behavior on H\textsc{eros}.
Our result reveals the blind spot of traditional STS benchmarks when evaluating SEs.\footnote{We release the dataset on \href{https://huggingface.co/datasets/dcml0714/Heros}{https://huggingface.co/datasets/dcml0714/Heros}.}

\end{abstract}

\begin{table}[th]
    \centering
    \begin{tabular}{c|ccccc}
        \toprule
         & \bf R1 & \bf R2 & \bf RL & \bf Lev & \bf Len\\
         \midrule
        STS-b & 55.8 & 32.5 & 53.2 & 0.54 & 12.2\\
        SICK-R & 61.2 & 37.4 & 56.2 & 0.53 & 10.0 \\
        H\textsc{eros} & 92.9 & 84.8 & 92.9 & 0.10& 13.8\\
        \bottomrule
    \end{tabular}
    \caption{
    We use the ROUGE F1 scores (R1, R2, RL) and the normalized Levenshtein distance (Lev) between sentence pairs to evaluate the degree of lexical overlaps in H\textsc{eros} and another two widely used STS benchmarks.
    A higher ROUGE score and a lower normalized Levenshtein distance imply higher lexical overlaps. 
    Len is the average sentence length.
    Please find details about the metrics used here in Appendix~\ref{app: Comparing the Lexical Overlaps of Different Datasets}.}
    \label{tab:statstic}
\end{table}

\section{Introduction}
Sentence encoders (SEs) are fundamental building blocks in miscellaneous natural language processing (NLP) tasks, including natural language inference, paraphrase identification, and retrieval~\citep{gillick2018end, lan2018neural}.
SEs are mostly evaluated with the semantic textual similarity (STS) datasets~\citep{agirre-etal-2016-semeval, cer-etal-2017-semeval} and SICK-R~\citep{marelli2014sick}, which consist of sentence pairs with human-labeled similarity scores.
The performance of the SEs is summarized using Spearman’s correlation coefficient between the human-labeled similarity and the cosine similarity obtained from the SE.

While the STS benchmarks are widely adopted, there are two problems with these benchmarks.
First, the performance on the STS dataset does not reveal much about what kind of sentence pairs would the SE deem \textit{similar}.
Spearman's correlation coefficient only tells us how correlated the sentence embedding cosine similarity and the ground truth similarity are.
However, the idea of what is similar can vary among different people and depend on the task at hand. 
Therefore, just because the sentence embedding cosine similarity is strongly correlated to the ground truth similarity, it does not provide much information about the specific type of similarity that the SE captures.
Prior works mostly resort to a few hand-picked examples to illustrate what kind of sentence pairs an SE would consider similar or dissimilar~\citep{gao2021simcse,chuang-etal-2022-diffcse,wang2022sncse}.
But it is hard to fully understand the traits of an SE by using only a few hand-picked samples.

The second issue is that sentence pairs in the STS-related benchmarks often have low lexical overlaps, as shown in Table~\ref{tab:statstic}, making it unclear how the SEs will perform on sentence pairs with high lexical overlaps, which exist in real-world applications such as adversarial attacks in NLP. 
Adversarial samples in NLP are constructed by replacing some words in an original sentence with some other words~\citep{alzantot-etal-2018-generating}, and the original sentence and the adversarial sample will have high lexical overlaps.
SEs are often adopted to check the semantic similarity between the original sentence and the adversarial sample~\citep{garg-ramakrishnan-2020-bae,li-etal-2020-bert-attack}.
If we do not know how SEs perform on high lexical overlap sentences, using them to check semantic similarity is meaningless.

To address the above issues, we construct and release a new dataset, H\textsc{eros}: \underline{\textbf{H}}igh-l\underline{\textbf{e}}xical ove\underline{\textbf{r}}lap diagn\underline{\textbf{o}}stic dataset for \underline{\textbf{s}}entence encoders, for evaluating SEs.
H\textsc{eros} is composed of six subsets, and each subset includes 1000 sentence pairs with very high lexical overlaps.
For the two sentences in a sentence pair, one of them is created by modifying the other sentence based on certain rules, and each subset adopts a different rule.
These rules are (1) replacing a word with a synonym, (2) replacing a word with an antonym, (3) replacing a word with a random word, (4) replacing a word with its typo, and (5,6) negating the sentence.
By comparing the sentence embedding cosine similarity of sentence pairs in different subsets, we can understand what kind of sentence pairs, when they have high lexical overlaps, would be considered similar by an SE.
We evaluate 60 sentence embedding models on H\textsc{eros} and reveal many intriguing and unreported observations on these SEs.

While some prior works also crafted sentence pairs to understand the performance of SEs, they either do not make the datasets publicly available~\citep{zhu2018exploring,zhu-de-melo-2020-sentence} or do not consider so many SEs as our paper does~\citep{barancikova-bojar-2020-costra}, especially unsupervised SEs.
Our contribution is relevant and significant as it provides a detailed understanding of SEs using a new dataset.
The contribution and findings of this paper are summarized as follows:
\begin{itemize}
    \item We release H\textsc{eros}, a high-quality dataset consisting of 6000 sentence pairs with high lexical overlaps.
    H\textsc{eros} allows researchers to systematically evaluate what sentence pairs would be considered similar by SEs when the lexical overlap is high.

    \item We evaluate 60 SEs on H\textsc{eros} and reveal several facts that were never reported before or only studied using a few hand-picked examples.

    \item We show that supervised SEs trained for different downstream tasks behave differently on different subsets of H\textsc{eros}, indicating that the SEs for different tasks encode different concepts of similarity.

    \item We find that all unsupervised SEs are considerably insensitive to negation, and further fine-tuning on NLI datasets makes them acquire the concept of negation. 

    \item We observe that SEs can have very different performances on different subsets of H\textsc{eros} even if their average STS benchmark performance difference is less than 0.2 points.
    
\end{itemize}


\section{H\textsc{eros}}
H\textsc{eros} consists of six subsets, and each subset consists of 1000 sentence pairs. 
The six subsets are \textbf{Synonym}, \textbf{Antonym}, \textbf{Typo}, \textbf{Random MLM}, and two types of \textbf{Negation} subsets.
In all subsets, each pair of sentences have high lexical overlap, and the two sentences only differ in at most one content word; we call these paired sentences \textit{minimal pairs}.

The dataset is constructed from the GoEmotions dataset~\citep{demszky2020goemotions}, a dataset for emotion classification collected from Reddit comments. 
We select one thousand sentences from GoEmotions and replace \textit{one word} in the original sentences with its synonym, antonym, a typo of the replaced word, and a random word obtained from BERT~\citep{devlin-etal-2019-bert} mask-infilling.
Last, we convert the original sentence into its negation using two different rules. 
After this process, we obtain six sentences for each of the 1000 selected sentences.
We pair the \textbf{original sentence} and a \textbf{converted sentence} to form a minimal pair, which has high lexical overlaps.
We will explain the above process in detail in Section~\ref{subsection: Dataset construction}.
Samples from H\textsc{eros} are shown in Table~\ref{tab: hero examples}.

\subsection{Motivation and Intended Usage}
Unlike traditional STS benchmark datasets that asked humans to assign a similarity score as the ground truth similarity, H\textsc{eros} does not provide a ground truth similarity score for sentence pairs.
This is because it is difficult to define how similar two sentences should be for them to be given a certain similarity score.
Moreover, the concept of similarity differs in downstream tasks. 
For example, in paraphrase tasks, a Negation minimal pair is considered semantically different; but for a retrieval task, we might consider them similar.

Thus, instead of providing a ground truth label for each sentence pair and letting future researchers pursue state-of-the-art results on H\textsc{eros}, we hope this dataset is used for diagnosing the characteristics of an SE.
Specifically, one can compare the average sentence embedding cosine similarity of sentence pairs in different subsets to understand what kind of similarity is captured by the sentence embedding model.

Different subsets in H\textsc{eros} capture various aspects of semantics.
Comparing the average cosine similarity between minimal pairs in Synonym and Antonym allows one to understand whether replacing a word with an antonym is more dissimilar to the original semantics than replacing a word with a synonym.
The average cosine similarity between minimal pairs in Negation can tell us how negation affects sentence embedding similarity.
Typos are realistic and happen every day.
While humans can infer the original word from a typo and get the original meaning of the sentence, it will be interesting to see how the typos affect the sentences' similarity with the original sentences.
The Random MLM subset can tell us how similar the sentence embedding can be when two sentences are semantically different but with high lexical overlaps.
By comparing the performance of different SEs on different subsets in H\textsc{eros}, we can further understand the trait of different SEs.

\begin{table*}[t]
    \centering
    \begin{tabular}{p{0.10\linewidth}p{0.4\linewidth}p{0.4\linewidth}}
    \toprule
       \bf Subset & \bf Example (adjective) & \bf Example (verb) \\
       \midrule
       Original & \it And that is why it is (or was) \textcolor{NavyBlue}{\textbf{illegal}}. & \it You do not know how much that \textcolor{NavyBlue}{\textbf{boosted}} my self-esteem right now. \\
       \midrule
       Synonym & \it And that is why it is (or was) \textcolor{OliveGreen}{\textbf{illegitimate}}. & \it You do not know how much that \textcolor{OliveGreen}{\textbf{increased}} my self-esteem right now.\\
       \midrule
       Antonym & \it And that is why it is (or was) \textcolor{OliveGreen}{\textbf{legal}}. & \it You do not know how much that \textcolor{OliveGreen}{\textbf{lowered}} my self-esteem right now.\\
       \midrule
       Random MLM & \it And that is why it is (or was) \textcolor{OliveGreen}{\textbf{here}}. & \it  You do not know how much that \textcolor{OliveGreen}{\textbf{affects}} my self-esteem right now.\\
       \midrule
       Typo & \it And that is why it is (or was) \textcolor{OliveGreen}{\textbf{illiegal}}. & \it You do not know how much that \textcolor{OliveGreen}{\textbf{booste}} my self-esteem right now.\\
       \midrule
       Negation (Main) & \it And that is \textcolor{OliveGreen}{\textbf{not}} why it is (or was) illegal. & \it You \textcolor{OliveGreen}{\textbf{do}} know how much that boosted my self-esteem right now.\\
       \midrule
       Negation (Antonym) & \it And that is why it is (or was) \textcolor{OliveGreen}{\textbf{not}} illegal. & \it You do not know how much that \textcolor{OliveGreen}{\textbf{did not boost}} my self-esteem right now.\\
    \bottomrule
    \end{tabular}
    \caption{Two examples from H\textsc{eros}. 
    One example selects a verb while the other selects an adjective for replacement.
    The first row shows the original sentences in the GoEmotions, and the words highlighted in \textcolor{NavyBlue}{\textbf{blue}} are the words to be replaced.
    Starting from the second rows are the corresponding sentences obtained from the original sentence for different subsets, and the changes compared with the original sentence are highlighted in \textcolor{OliveGreen}{\textbf{green}}.
    }
    \label{tab: hero examples}
\end{table*}

\subsection{Dataset construction}
\label{subsection: Dataset construction}

\subsubsection{Raw dataset preprocessing}
\label{subsubsection: Raw dataset preprocessing}
H\textsc{eros} are constructed from GoEmotions.
For the sentences in GoEmotions, we only select sentences whose lengths are more than 8 words and less than 25 words. 
We filter out sentences with cursing, and we use \href{https://pypi.org/project/language-tool-python/}{language-tool} to filter out sentences that language-tool find ungrammatical.
We manually remove the sentences that we find offensive or harmful.
The selected sentences are called \textbf{original sentences} in our paper.
More details on preprocessing are presented in Appendix~\ref{appendix: Dataset Preprocessing: Anonomization}.

\subsubsection{Selecting which word to replace}

The next step is to determine which word to replace in the original sentences obtained from preprocessing. 
The selected word must be (1) semantically significant to the original sentence so that when it is replaced with a non-synonym word, the two sentences will have vastly different meanings and would be considered contradictory in an NLI task.
(2) The selected word must have synonyms and antonyms at the same time since it will be replaced with its synonyms and antonyms. 
We only select verbs and adjectives for replacement because changing them greatly alters the semantics of a sentence. 
Sentences that do not contain a word that satisfies the two criteria are dropped.

\subsubsection{Synonym and Antonym Subsets}
\label{subsection: Synonym and Antonym}
The first subset in H\textsc{eros} includes the minimal pairs formed by replacing a word in the original sentence with its synonym; the second subset includes the minimal pairs formed by replacing a word with its antonym.
After selecting the word to be replaced, we determine what synonym and antonym should be used for replacement.
There are three principles for replacement: (1) the replacement should fit in the context, (2) the synonym should match the word sense of the original word in the sentence\footnote{A word can have different word senses, and each word sense has its own synonym sets. Synonym sets of different word senses might be different.}, and (3) the collocating words (e.g., prepositions, definite articles) may also need to be modified.
The three guiding principles make this process require high proficiency in English and this process is impossible to be done using an automatic process.
Thus, this process is performed by the authors ourselves.
We use our proficiency in English and four online dictionaries to select the replacement words.
The four resources are \href{https://www.thesaurus.com/}{thesaurus.com}, \href{https://www.merriam-webster.com/}{thesaurus of Merriam-Webster}, \href{https://www.freecollocation.com/}{Online Oxford Collocation Dictionary}, and \href{https://dictionary.cambridge.org/}{Cambridge Dictionary}.
This step takes 72 hours.

\subsubsection{Random MLM}
\label{subsection: Random MLM}
The third subset in H\textsc{eros} is obtained by replacing the word to be replaced with a random word predicted by a masked language model.
We mask the word to be replaced by \texttt{[MASK]} token and use \href{https://huggingface.co/bert-large-uncased}{\texttt{bert-large-uncased}} to fill in the masked position.
We filter out the synonym, antonym, and their derivational forms\footnote{For example, different tenses of a verb.} from the masked prediction using WordNet and \href{https://github.com/bjascob/LemmInflect}{LemmInflect}. 
Additionally, we filter out punctuations and subword tokens that are not complete words.
Moreover, we manually filter out mask predictions that are very similar in meaning to the original word when used in the same context.
This is because even if a word is not a synonym of the word to be replaced, it may still express the same meaning when used in the same context.
For example, "great" is not a synonym of "good" according to \href{http://wordnetweb.princeton.edu/perl/webwn?s=great&sub=Search+WordNet&o2=&o0=1&o8=1&o1=1&o7=&o5=&o9=&o6=&o3=&o4=&h=}{WordNet}, but their meaning is very similar. 
The resulting sentences can be ungrammatical in very few cases, but we leave them as is.

\subsubsection{Typos}
\label{subsection: Typos}
The fourth subset in H\textsc{eros} is constructed by swapping the word to be replaced with its typo.
Typos are spelling or typing errors that occur in real life.
If the word to be typoed is in the \href{https://en.wikipedia.org/wiki/Wikipedia:Lists_of_common_misspellings/For_machines}{Wikipedia lists of common misspellings}, we replace the word with its typo in the list.
If the word is not in the common misspelling list, we create a typo by randomly deleting or replacing one character or swapping two different characters in the word~\citep{he2021model}. 

\subsubsection{Negation}
\label{subsection: Negation}
The last subset in H\textsc{eros} is constructed by negating the original sentence.
Negation can happen at different levels in a sentence, and we create two different types of negation datasets based on where the negation happens.
The first one is negating the main verb, which is the action performed by the subject, in the sentence.\footnote{When selecting the original sentences, we do not consider interrogative sentences, so we will not create a negative interrogative sentence. An interrogative sentence and its negation ask the same question and are not semantically different.}
If the main verb is not negated, we negate it by adding appropriate auxiliary verbs and the word "not".
If the main verb is already negated, we directly remove the word "not" and do not remove the auxiliary verb.
We call this type of negation dataset the \textbf{Negation (Main)}.

The other type of negation dataset is related to the Antonym subset.
A minimal pair in the Antonym subset is formed by replacing a word in a sentence with its antonym.
This \textit{implicitly} negates the meaning of the original sentence.
Here, we construct another subset called \textbf{Negation (Antonym)}, which \textit{explicitly} negates the word that is replaced with an antonym in the Antonym subset.
Given a sentence pair in the Antonym dataset, there is a verb or adjective in the original sentence that is replaced by its antonym in the converted sentence.
We directly negate that word in the original sentence by adding "not" in front of an adjective or adding "not" and an auxiliary verb for verbs.\footnote{If the negation of the minimal pair in the Antonym dataset happens at the main verb, the sentence pair in Negation (Main) and Negation (Antonym) will be the same.}.
If the word is already negated, we remove the "not". 
These sentences might sound a bit strange, but they are still understandable. 
This type of negation dataset is called Negation (Antonym) because the negation is in the same place as the antonym replacement in the Antonym subset.

\begin{figure*}[ht!]
\centering
\includegraphics[clip, trim = 20px 220px 20px 20px,width=0.95\linewidth]{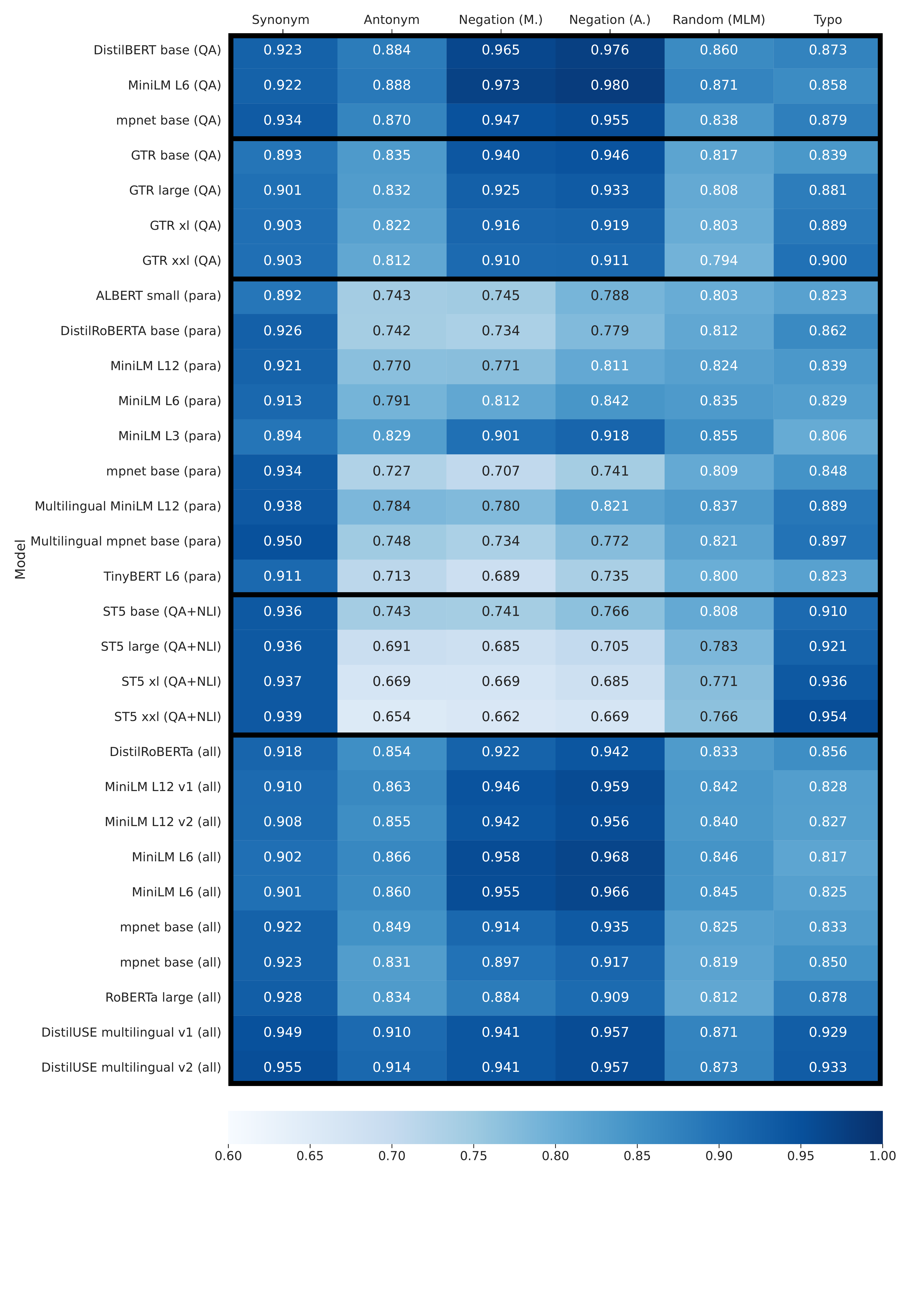}
\caption{Normalized cosine similarity of supervised SEs. 
We group SEs that use different training datasets or training procedures together.
We denote the datasets used to train the SEs in parentheses.
}
\label{fig:sup.pdf}
\end{figure*}

\section{Comparing 60 Sentence Embedding Models}
\label{section: Benchmarking 60 Sentence Embedding Models}
In this section, we compare the behavior of 60 SEs on H\textsc{eros}. 
Detailed information about the SEs, including training data and model size, is listed in Appendix~\ref{app: Sentence Encoders Used}. 
We calculate the cosine similarity between each minimal pair in H\textsc{eros} and normalize it by a baseline cosine similarity to remove the effect of anisotropic embedding space~\citep{ethayarajh-2019-contextual,li-etal-2020-sentence}.
The baseline cosine similarity is calculated by averaging the similarity between 250K random sentence pairs (details in Appendix~\ref{app:normalize}). 
We report the average normalized similarity of different subsets in H\textsc{eros} for each SE. For simplicity, we will use "similarity" to refer to the normalized cosine similarity.

\subsection{Supervised SEs}
We use 30 supervised transformer-based SEs in the \href{https://www.sbert.net/}{SentenceTransformers} toolkit~\citep{reimers-2019-sentence-bert,reimers-2020-multilingual-sentence-bert}.
These SEs are trained supervisedly using different datasets for specific downstream tasks.
The results are presented in Figure~\ref{fig:sup.pdf}.
In Figure~\ref{fig:sup.pdf}, we group SEs into groups based on what training dataset they used.
We denote the datasets used to fine-tune the SEs in the parentheses in Figure~\ref{fig:sup.pdf}.
There are a lot of interesting observations one can obtain from Figure~\ref{fig:sup.pdf}, and we are just listing some of those observations.

\textbf{SEs fine-tuned only on QA datasets are insensitive to negation:}
The first and second blocks in Figure~\ref{fig:sup.pdf} include different SEs obtained from fine-tuning on QA datasets using contrastive learning~\citep{hadsell2006dimensionality}.
In the fine-tuning stage, a positive pair for contrastive learning is a pair of question and the answer to the question.
The high similarity of the two Negation subsets can be explained by the dataset type used for fine-tuning: whether the answer is negated or not, it may still be considered a valid answer to the question.
Hence, it is reasonable that a sentence and its negation have high similarity.
We also find that replacing a word with a typo will cause the resulting sentence to have lower similarity with the original sentence compared with replacing the word with a synonym.
While humans can understand the real meaning of a typo word, this is not the case for the SEs.

\begin{figure*}[ht!]
\centering
\includegraphics[clip, trim = 20px 220px 20px 20px,width=1.0\linewidth]{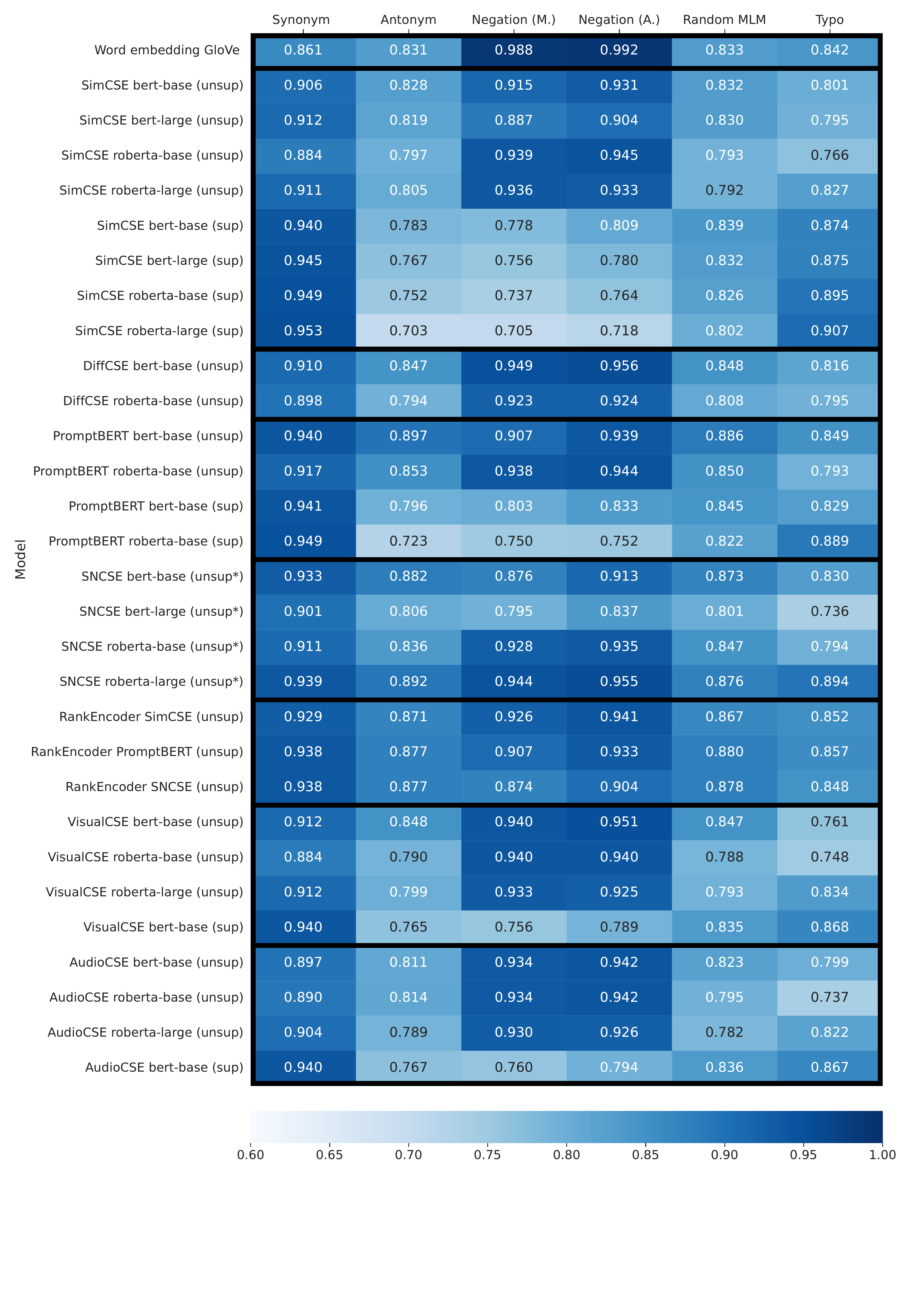}
\caption{Normalized cosine similarity of unsupervised SEs and their derived supervised SEs.
We group the SEs based on the unsupervised fine-tuning method.
}
\label{fig:unsup.pdf}
\end{figure*}

\textbf{SEs fine-tuned from T5 are less sensitive to typos when the model size scales up:}
The GTR models~\citep{ni2021large} in the second block of Figure~\ref{fig:sup.pdf} and the ST5 models~\citep{ni-etal-2022-sentence} in the fourth block are SEs fine-tuned from T5~\citep{raffel2020exploring}.
Although the two types of models are trained using different datasets, we find that their performance on the Typo subset shares an interesting trend when the model size scales up from the smallest base-size model to the largest xxl-size model:
The similarity on the Typo subset grows higher as the model gets larger and can be as high as or higher than the similarity of the Synonym subset; meanwhile, the similarity on the Synonym subset is almost unchanged when the model size gets larger.
This shows that deeper model can better mitigate the negative impact of typos on sentence embeddings.

\textbf{SEs fine-tuned on paraphrase datasets are extremely sensitive to negations and antonyms:}
The third block in Figure~\ref{fig:sup.pdf} includes the results of SEs fine-tuned on paraphrase datasets using contrastive learning.
Paraphrase datasets include a combination of different datasets such as premise-hypothesis pairs in NLI datasets and duplicate question pairs.
Contrary to the previous paragraph which shows fine-tuning only using question-answer pairs makes the model insensitive to negation, we see a completely different result in the third block of Figure~\ref{fig:sup.pdf}.
We infer that this is mainly due to the NLI datasets used for fine-tuning: negating the original sentence results in a sentence that semantically contradicts the original sentence, and will be considered as a hard negative in contrastive learning.
Hence, SEs fine-tuned on NLI will be very sensitive negation.
For the same reason, these SEs are also sensitive to replacing words with antonyms.
The only exception is the \texttt{MiniLM L3 (para)} model~\citep{wang2020minilm}, which has very high similarity on the Negation subsets and is even higher than the Synonym subset.
We hypothesize that this is because the number of parameters of the model and the sentence embedding dimension are too small, thus limiting the expressiveness of the sentence embeddings.

\textbf{SEs fine-tuned on all available sentence pair datasets are again insensitive to negations:}
The models in the last block in Figure~\ref{fig:sup.pdf} are fine-tuned on all available sentence-pair training data, denoted as \textit{(all)}.
The training data consist of 32 datasets and have a total of 1.17B sentence pairs, including question-answer pairs in QA datasets, premise-hypothesis pairs in NLI dataset, and context-passage pairs in retrieval datasets. 
In the fifth block, the similarity between sentence pairs from the two Negation subsets is very high and is even higher than the similarity of the Synonym subset for most models.
This means that when using these models for retrieval, given a source sentence, it is more likely to retrieve the negation of the source sentence, instead of another sentence that only differs from the source sentence by a synonym.
While these models are also fine-tuned on NLI datasets, the NLI datasets only compose 0.24\% of the whole training data.
This makes the models in this block much less sensitive to negations, compared with models fine-tuned mainly with NLI datasets (e.g., ST5) and models fine-tuned on paraphrase datasets. 


\textbf{H\textsc{eros} reveal different characteristics of different SEs:}
Overall, we see that even if the sentences in H\textsc{eros} all have high lexical overlaps, the similarity score can still be very different among H\textsc{eros} subsets for the same SE.
H\textsc{eros} also shows that how the concept of similarity is encoded by an SE is highly related to what the SE is trained on.
This further allows us to understand what kind of similarity is required by the task related to the training dataset.
For example, NLI tasks consider negation pairs as dissimilar while question-answer pair retrieval task considers negation to be similar.
Such interesting observations are not revealed by any prior SE benchmarks, making H\textsc{eros} very valuable.
It will also be interesting to see if there is any correlation between an SE's performance on different subsets in H\textsc{eros} and different downstream tasks in SentEval~\citep{conneau-kiela-2018-senteval}; we save this in future work.

\subsection{Unsupervised SEs}
Next, we turn our attention to unsupervised SEs.
Unlike supervised SEs that are fine-tuned on labeled pairs of sentences, unsupervised SEs are trained using specially designed methods that do not use labeled sentence pairs.
Most of these unsupervised SEs can be further fine-tuned on NLI datasets to further improve the performance on the STS benchmarks~\cite{gao2021simcse, jiang2022promptbert,jian-etal-2022-non}.
We show the result on H\textsc{eros} of 7 different types of unsupervised SEs and their derived supervised SEs in Figure~\ref{fig:unsup.pdf}.

For the completeness of the result, we also report the performance of sentence embeddings calculated by averaging the GLoVe embeddings~\citep{pennington-etal-2014-glove} in the sentence.
The result is presented in the first row in Figure~\ref{fig:unsup.pdf}.
We observe that the sentence before and after negation have very high similarity, and the similarity is much higher than replacing one word with its synonym or antonym.
This shows that negation words have a very small contribution to the sentence embedding obtained from averaging the GLoVe embeddings.

\textbf{Unsupervised SEs are insensitive to negation:}
Unsupervised SEs, denoted with \textit{unsup} in Figure~\ref{fig:unsup.pdf}, have high similarity on Negation subsets, sometimes even higher than Synonym subsets. 
SNCSE~\citep{wang2022sncse} models are an exception, where Negation subsets may have a lower similarity. 
SNCSE uses the dependency tree of a sentence to convert it into its negation as a "soft-negative" in contrastive learning, but it needs a dependency parser, making it not truly unsupervised. 
Hence, we use \textit{unsup*} to denote SNCSE models in Figure~\ref{fig:unsup.pdf}.
The lower similarity on Negation datasets is not consistent for different SNCSE models, possibly due to a poor negation method in the \href{https://github.com/Sense-GVT/SNCSE/blob/main/generate_soft_negative_samples.py}{implementation} of SNCSE that does not consider negative contractions, resulting in low-quality augmented data.

\textbf{Further supervised fine-tuning on NLI datasets significantly change the model's behavior on H\textsc{eros}:}
Fine-tuning unsupervised SEs on NLI datasets (denoted with \textit{sup} in Figure~\ref{fig:unsup.pdf}) leads to a significant drop in similarity on Negation and Antonym and an increase in similarity on the Synonym subset. 
This show that supervised fine-tuning greatly changes how SEs encode similarity. 
An interesting trend is that after fine-tuning, similarity on the Typo subset increases for most models, likely because the SE better captures semantic similarity and pays less attention to superficial lexical form.

\textbf{Almost all SEs rate Negation (Main) to be less similar compared with Negation (Antonym)}
Recall that the Negation (Main) subsets are created by negating the \textbf{main} verb while the Negation (Antonym) subset does not always negate the main verb.
The lower similarity on the Negation (Main) subset shows that SEs consider negating the main verb to be less similar to the original sentence, compared with negating other positions in the original sentence.
This implies that the SEs can capture the level of the verb in the dependency tree of the sentence, and it considers negating the main verb to be more influential to sentence embeddings.

\textbf{Close performance on STS benchmarks can have different behaviors on H\textsc{eros}:}
We find that two SEs that achieve similar average performance on STS benchmarks (STS 12-17, STS-b, and SICK-R) can perform very differently on H\textsc{eros}.
For example, \texttt{RoBERTa large (all)} and \texttt{DistilRoBERTa base (para)} in Figure~\ref{fig:sup.pdf} have similar average STS scores ($81.07$ and $81.12$, respectively), but the former have very high similarity on the Negation subsets while the latter does not. 
This is also the case for \texttt{SNCSE roberta-large} in Figure~\ref{fig:unsup.pdf} and \texttt{mpnet base (para)} in Figure~\ref{fig:sup.pdf}, which have average scores of $81.77$ and $81.57$ on the STS benchmarks, respectively.
This shows that H\textsc{eros} can reveal some traits of the SEs that the traditional STS benchmarks cannot identify.

\section{Conclusion}

We introduce H\textsc{eros}, a new dataset of 6000 human-constructed sentence pairs with high lexical overlaps.
It is composed of 6 subsets that capture different linguistic phenomena. 
Evaluating an SE on H\textsc{eros} can reveal what kind of sentence pairs the SE considers similar. 
H\textsc{eros} fills a void in current SE evaluation methods, which only use correlation coefficients with human ratings or performance on downstream tasks to summarize an SE, and mainly use sentence pairs with low lexical overlaps.
We use H\textsc{eros} to evaluate 60 models and reveal numerous new observations. 
We believe that H\textsc{eros} can aid in interpreting SE behavior and comparing the performance of different SEs.

\section*{Limitations}
The SEs in this paper are mainly transformer-based SEs, and we are not sure whether the observations hold for other SEs.
However, considering that transformer-based SEs dominate the current NLP community, we think it is fine to only evaluate 59 transformer-based SEs.
Another limitation is that the sentences in H\textsc{eros} are converted from Reddit, which is an online forum and the texts on Reddit may be more casual and informal.
This makes the sentence pairs in H\textsc{eros} tend to be more informal.
Users should note such a characteristic of the sentence pairs of H\textsc{eros} and beware that the results obtained using H\textsc{eros} may be different from the results obtained using more formal texts.
An additional limitation is that there can be more diverse rules to create different sentence pairs other than the six subsets included in H\textsc{eros}, and our paper cannot include them all.
As a last limitation, during the construction of H\textsc{eros}, we remove sentences that are ungrammatical based on \href{https://pypi.org/project/language-tool-python/}{language-tool}, so our results may not generalize to ungrammatical sentences.

\section*{Ethics Statement}
The main ethical concern in this paper is our dataset, H\textsc{eros}.
H\textsc{eros} is constructed from an existing dataset, GoEmotion.
As listed in the \href{https://github.com/google-research/google-research/blob/master/goemotions/goemotions_model_card.pdf}{model card of GoEmotion}, GoEmotion contains biases in Reddit and some offensive contents.
As stated in Section~\ref{subsubsection: Raw dataset preprocessing}, the authors have tried our best to remove all content that we find to be possibly offensive to users.
We cannot guarantee that our standard of unbiased and unharmful fits everyone.
Thus, we also remind future users of H\textsc{eros} to be aware of such possible harms.
To make sure the accessibility of our paper, we have used an \href{https://www.color-blindness.com/coblis-color-blindness-simulator/}{online resource} to carefully check that the figures in the paper are interpretable for readers of different backgrounds.

\section*{Acknowledgements}
We thank Yeon Seonwoo for providing the pretrained checkpoints of RankEncoder~\citep{seonwoo2022ranking}.

\bibliography{custom}
\bibliographystyle{acl_natbib}

\appendix

\section{Further Details of H\textsc{eros}}
\subsection{Dataset Preprocess}
\label{appendix: Dataset Preprocessing: Anonomization}
The sentences in GoEmotions are already anonymized, where the names of people are replaced with a special \texttt{[NAME]} token, so we do not need to further perform anonymization. 
We filter out all sentences that have more than one \texttt{[NAME]} token and replace all \texttt{[NAME]} with a gender-neutral name "Jackie".

\subsection{Dataset License}
H\textsc{eros} is constructed based on the GoEmotion dataset~\citep{demszky2020goemotions}.
GoEmotion is released under the Apache 2.0 license, so our modification and redistribution to GoEmotion are granted by the dataset license.
\textbf{Our dataset, H\textsc{eros} is also released under the \href{https://www.apache.org/licenses/LICENSE-2.0}{Apache 2.0 license.}}

\section{Comparing the Lexical Overlaps of Different Datasets}
\label{app: Comparing the Lexical Overlaps of Different Datasets}
In Table~\ref{tab:statstic}, we show the basic statistics of three different datasets.
We use the ROUGE F1 and Levenshtein distance to quantify the lexical overlap between sentence pairs of a dataset.
The statistics of H\textsc{eros} is averaged over different subsets, and those of STS-b and SICK-R are calculated based on the test set.

R1, R2, and RL: ROUGE F1 score between the sentence pairs.
(R1 and R2: unigram and bigram overlap; RL: longest common subsequence.)
We use the implementation of \href{https://pypi.org/project/rouge/}{python rouge 1.0.1} to calculate the ROUGE score.

Lev is the average normalized \textbf{token-level} Levenshtein distance among the sentence pairs, and the normalized Levenshtein distance is the Levenshtein distance between two sentences divided by the length of the longer sequence of the sentence pairs.
We first tokenize the sentence using the \href{https://huggingface.co/bert-base-uncased}{tokenizer of \texttt{bert-base-uncased}} and calculate the Levenshtein distance between the token ids of the sentence pairs.
We normalize the Levenshtein distance to make it falls in the range of $[0,1]$.

The average sentence length is the average number of tokens per sentence, and the tokens are obtained by using the \href{https://huggingface.co/bert-base-uncased}{tokenizer of \texttt{bert-base-uncased}}.

\section{Supplementary Materials for Sentence Encoders}
\label{app: Sentence Encoders Used}

\subsection{Supervised Sentence Encoders}
\label{app: sup SE}
Table~\ref{tab:SEs} shows the number of parameters and the sentence embedding dimension of the SEs used in this paper.
\subsubsection{Datasets Used to Train Supervised SEs}
The datasets indicated in Figure~\ref{fig:sup.pdf} is listed as follows:
\paragraph{all}
        Reddit comments (2015-2018)~\citep{henderson-etal-2019-repository}, 
        S2ORC Citation pairs (Abstracts)~\citep{lo-etal-2020-s2orc},
        WikiAnswers Duplicate question pairs~\citep{10.1145/2623330.2623677},
        PAQ (Question, Answer) pairs~\citep{lewis-etal-2021-paq},
        S2ORC Citation pairs (Titles)~\citep{lo-etal-2020-s2orc},
        S2ORC (Title, Abstract)~\citep{lo-etal-2020-s2orc},
        \href{https://huggingface.co/datasets/flax-sentence-embeddings/stackexchange_xml}{Stack Exchange (Title, Body) pairs},
        MS MARCO triplets~\citep{10.1145/3404835.3462804},
        GOOAQ: Open Question Answering with Diverse Answer Types~\citep{gooaq2021},
        Yahoo Answers (Title, Answer)~\citep{zhang2015character},
        \href{https://huggingface.co/datasets/code_search_net}{Code Search},
        COCO Image captions~\citep{lin2014microsoft},
        SPECTER citation triplets~\citep{cohan-etal-2020-specter},
        Yahoo Answers (Question, Answer)~\citep{zhang2015character},
        Yahoo Answers (Title, Question)~\citep{zhang2015character},
        SearchQA~\citep{dunn2017searchqa},
        Eli5~\citep{fan-etal-2019-eli5},
        Flickr 30k~\citep{young2014image},
        \href{https://huggingface.co/datasets/flax-sentence-embeddings/stackexchange_xml}{Stack Exchange Duplicate questions (titles)},
        SNLI~\citep{bowman-etal-2015-large},
        MNLI~\citep{williams-etal-2018-broad},
        \href{https://huggingface.co/datasets/flax-sentence-embeddings/stackexchange_xml}{Stack Exchange} Duplicate questions (bodies),
        \href{https://huggingface.co/datasets/flax-sentence-embeddings/stackexchange_xml}{Stack Exchange} Duplicate questions (titles+bodies),
        Sentence Compression~\citep{filippova-altun-2013-overcoming},
        Wikihow~\citep{koupaee2018wikihow},
        Altlex~\citep{hidey-mckeown-2016-identifying},
        Quora Question Triplets~\citep{wang2018glue},
        Simple Wikipedia~\citep{coster-kauchak-2011-simple},
        Natural Questions (NQ)~\citep{kwiatkowski2019natural},
        SQuAD2.0~\citep{rajpurkar-etal-2016-squad}, and
        \href{https://huggingface.co/datasets/trivia_qa}{TriviaQA}.

\paragraph{QA}
All the QA datasets in \textbf{\textit{all}}.

\paragraph{paraphrase}
SNLI~\citep{bowman-etal-2015-large},
MNLI~\citep{williams-etal-2018-broad},
Simple Wikipedia~\citep{coster-kauchak-2011-simple},
Altlex~\citep{hidey-mckeown-2016-identifying},
MS MARCO triplets~\citep{10.1145/3404835.3462804},
Quora Question Triplets~\citep{wang2018glue},
COCO Image captions~\citep{lin2014microsoft},
Flickr 30k~\citep{young2014image},
Yahoo Answers (Title, Question)~\citep{zhang2015character},
\href{https://huggingface.co/datasets/flax-sentence-embeddings/stackexchange_xml}{Stack Exchange} Duplicate questions (titles+bodies) and 
WikiAtomicEdits~\citep{faruqui-etal-2018-wikiatomicedits}.

\paragraph{GTR fine-tuning data: QA+MRC}
Natural Questions (NQ)~\citep{kwiatkowski2019natural}, MS MARCO triplets~\citep{10.1145/3404835.3462804}, input-response pairs and question-answer pairs from online forums and QA websites including Reddit, Stack-Overflow, etc.\footnote{\citet{ni2021large} does not specify the exact online forums.}

\paragraph{ST5 fine-tuning data: QA+NLI}
SNLI~\citep{bowman-etal-2015-large} and question-answer pairs from community QA websites.

\begin{table*}[t!]
    \centering
    \begin{tabular}{ccc}
    \hline
        Model and Link & \#Param & $d_{emb}$ \\
        \hline
        \href{https://huggingface.co/sentence-transformers/average_word_embeddings_glove.6B.300d}{Word embedding GloVe} & 120M & 300 \\
\href{https://huggingface.co/sentence-transformers/multi-qa-distilbert-cos-v1}{DistilBERT base (multi-QA)} & 66M & 768 \\
\href{https://huggingface.co/sentence-transformers/multi-qa-MiniLM-L6-cos-v1}{MiniLM L6 (multi-QA)} & 22M & 384 \\
\href{https://huggingface.co/sentence-transformers/multi-qa-mpnet-base-cos-v1}{mpnet base (multiQA)} & 110M & 768 \\
\href{https://huggingface.co/sentence-transformers/paraphrase-albert-small-v2}{ALBERT small (paraphrase)} &  11M & 768 \\
\href{https://huggingface.co/sentence-transformers/paraphrase-distilroberta-base-v2}{DistilRoBERTA base v2 (paraphrase)} & 82M & 768\\
\href{https://huggingface.co/sentence-transformers/paraphrase-MiniLM-L12-v2}{MiniLM L12 v2 (paraphrase)} & 33M & 384 \\
\href{https://huggingface.co/sentence-transformers/paraphrase-MiniLM-L3-v2}{MiniLM L3 v2 (paraphrase)} & 17M & 384 \\
\href{https://huggingface.co/sentence-transformers/paraphrase-MiniLM-L6-v2}{MiniLM L6 v2 (paraphrase)} & 22M & 384 \\
\href{https://huggingface.co/sentence-transformers/paraphrase-mpnet-base-v2}{mpnet base v2 (paraphrase)} & 110M & 768  \\ 
\href{https://huggingface.co/sentence-transformers/paraphrase-multilingual-MiniLM-L12-v2}{Multilingual MiniLM L12 v2 (paraphrase)} & 33M & 384 \\
\href{https://huggingface.co/sentence-transformers/paraphrase-multilingual-mpnet-base-v2}{Multilingual mpnet base v2 (paraphrase)} & 110M & 768 \\
\href{https://huggingface.co/sentence-transformers/paraphrase-TinyBERT-L6-v2}{TinyBERT L6 v2 (paraphrase)} & 14.5M & 768 \\
\href{https://huggingface.co/sentence-transformers/gtr-t5-base}{GTR base} & 110M & 768  \\
\href{https://huggingface.co/sentence-transformers/gtr-t5-large}{GTR large} & 335M &768  \\
\href{https://huggingface.co/sentence-transformersgtr-t5-xl}{GTR xl} & 1,24B & 768 \\
\href{https://huggingface.co/sentence-transformers/gtr-t5-xxl }{GTR xxl} & 4.8B & 768  \\
\href{https://huggingface.co/sentence-transformers/sentence-t5-base}{Sentence-T5 base} & 110M & 768  \\
\href{https://huggingface.co/sentence-transformers/sentence-t5-large}{Sentence-T5 large} & 335M &768 \\
\href{https://huggingface.co/sentence-transformers/sentence-t5-xl}{Sentence-t5 xl} & 1,24B & 768   \\
\href{https://huggingface.co/sentence-transformers/sentence-t5-xxl}{Sentence-T5 xxl} & 4.8B & 768   \\
        \href{https://huggingface.co/sentence-transformers/all-distilroberta-v1}{DistilRoBERTa v1 (all)} & 82M & 768\\
        \href{https://huggingface.co/sentence-transformers/all-MiniLM-L12-v1}{MiniLM L12 v1 (all)}& 33M & 384\\
\href{https://huggingface.co/sentence-transformers/all-MiniLM-L12-v2}{MiniLM L12 v2 (all)} & 33M & 384  \\
\href{https://huggingface.co/sentence-transformers/all-MiniLM-L6-v1}{MiniLM L6 v1 (all)} & 22M & 384  \\
\href{https://huggingface.co/sentence-transformers/all-MiniLM-L6-v2}{MiniLM L6 v2 (all)} & 22M & 384 \\
\href{https://huggingface.co/sentence-transformers/all-mpnet-base-v1}{mpnet base v1 (all)} & 110M & 768 \\
\href{https://huggingface.co/sentence-transformers/all-mpnet-base-v2}{mpnet base v2 (all)} & 110M & 768  \\
\href{https://huggingface.co/sentence-transformers/all-roberta-large-v1}{RoBERTa large v1 (all)} & 355M & 1024 \\
\href{https://huggingface.co/sentence-transformers/distiluse-base-multilingual-cased-v1}{DistilUSE base multilingual v1 (all)} & 134M & 512  \\
\href{https://huggingface.co/sentence-transformersdistiluse-base-multilingual-cased-v2}{DistilUSE base multilingual v2 (all)} & 134M & 512  \\
\hline
    \end{tabular}
    \caption{The model sizes and embedding dimensions of the supervised SEs shown in Figure~\ref{fig:sup.pdf}. 
    The model names are clickable links.
    \# is the number of parameters of the SE, and $d_{emb}$ is the dimension of the sentence embedding.
    }
    \label{tab:SEs}
\end{table*}

\subsection{Unsupervised Sentence Encoders}
\label{app: unsup SEs}
The full list of unsupervised SEs and their supervised derivations we compared are: SimCSE~\citep{gao2021simcse}, DiffCSE~\citep{chuang-etal-2022-diffcse}, PromptBERT~\citep{jiang2022promptbert}, SNCSE~\citep{wang2022sncse}, RankEncoder~\citep{seonwoo2022ranking}, AudioCSE and VisualCSE~\citep{jian-etal-2022-non}.
For all the unsupervised SEs shown in Figure~\ref{fig:unsup.pdf}, if it is a base-size model, its number of parameters is roughly 110M; if it is a large-size model, its number of parameters is roughly 335M.
The \texttt{bert} models shown in Figure~\ref{fig:unsup.pdf} are all \texttt{uncased} models.

\section{Normalization}
\label{app:normalize}
For each SE, we first calculate the cosine similarity between each minimal pair in H\textsc{eros}.
However, if the embedding space is highly anisotropic~\citep{ethayarajh-2019-contextual,li-etal-2020-sentence}, the cosine similarity between two random sentences is expected to be rather high.
To remove the effect of anisotropic embedding space and better interpret the result, we normalize the cosine similarity by a baseline cosine similarity.
The baseline cosine similarity is calculated by the following procedure: 
We split the 1000 original sentences into the first 500 and the last 500 sentences, and 
calculate the average cosine similarity between the sentence embeddings of these 500 $\times$ 500 random sentence pairs.
This average cosine similarity, $\cos_{avg}$, gives us an idea of how similar sentence embedding can be for two randomly selected sentences.
Last, we normalize the cosine similarity of the minimal pairs to lessen the effect of anisotropy by the following formula:
\begin{equation}
    \cos_{normalized} = \frac{\cos_{orig} - \cos_{avg}}{1 - \cos_{avg}},
\end{equation}
where $\cos_{orig}$ is the original cosine similarity of a sentence pair and $\cos_{normalized}$ is the similarity after normalization.

\section{Runtime and Computation Resource}
The experiments on Section~\ref{section: Benchmarking 60 Sentence Embedding Models}, except T5 xl and xxl, are conducted on an NVIDIA 1080 Ti, and it takes less than one hour to run all the experiments.
The T5 xxl and xl models cannot be loaded on a 1080 Ti, and we use V100 to conduct the experiment of the SEs whose base models are T5 xl and xxl,
 which takes less than 15 minutes.

\end{document}